\documentclass[10pt,twocolumn,letterpaper]{article}

\usepackage[review,algorithms]{wacv}      

\usepackage{graphicx}
\usepackage{amsmath}
\usepackage{amssymb}
\usepackage{booktabs}

\usepackage{array}
\usepackage{makecell}

\usepackage[pagebackref,breaklinks,colorlinks]{hyperref}

\usepackage[capitalize]{cleveref}
\crefname{section}{Sec.}{Secs.}
\Crefname{section}{Section}{Sections}
\Crefname{table}{Table}{Tables}
\crefname{table}{Tab.}{Tabs.}
\usepackage{etoolbox}
\toggletrue{wacvfinal}

\def\confName{WACV}

\usepackage{graphicx}  

\begin{document}



\title{Enhanced Transformer-Based Tracking for Skiing Events: Overcoming Multi-Camera Challenges, Scale Variations and Rapid Motion - SkiTB Visual Tracking Challenge 2025
}

\author{
Akhil Penta, Vaibhav Adwani, Ankush Chopra \\
Tredence Analytics \\
{\tt\small \{akhil.penta, vaibhav.adwani, ankush.chopra\}@tredence.com}
}

\maketitle

\begin{abstract}
   Accurate skier tracking is essential for performance analysis, injury prevention, and optimizing training strategies in alpine sports. Traditional tracking methods often struggle with occlusions, dynamic movements, and varying environmental conditions, limiting their effectiveness. In this work, we used STARK \cite{stark} (Spatio-Temporal Transformer Network for Visual Tracking), a transformer-based model, to track skiers. We adapted STARK \cite{stark} to address domain-specific challenges such as camera movements, camera changes, occlusions, etc. by optimizing the model’s architecture and hyperparameters to better suit the dataset.

\end{abstract}
\section{Introduction}
\label{sec:intro}

Object tracking is a fundamental task in computer vision, involving the detection and tracking of moving objects across video frames. Modern tracking methods can be broadly categorized into three approaches: Detection-Based Tracking, Joint Detection and Tracking, and Attention/Transformer-Based Methods.
\textbf{Detection-Based Tracking (Tracking-by-Detection)} detects objects in each frame and associates them across frames. Algorithms like SORT \cite{sort} and DeepSORT \cite{deepsort} rely on Kalman Filters and deep appearance embeddings for better re-identification, while ByteTrack \cite{bytetrack} improves handling of low-confidence detections. This approach is modular and flexible but tends to be slower.
\textbf{Joint Detection and Tracking (End-to-End Tracking)} integrates detection and tracking within a single network, improving efficiency. Models like FairMOT \cite{fairmot} and JDE \cite{jde} jointly perform detection and feature extraction, reducing identity switches and handling occlusions well, but they struggle with long-range dependencies. \textbf{Attention/Transformer-Based Methods} leverage self-attention to capture long-range dependencies, excelling in occlusions and fast motions. Models such as TransTrack \cite{trans_track} and STARK \cite{stark} enhance tracking robustness but are computationally expensive and struggle with abrupt camera transitions in multi-camera footage.

\section{Data Overview}
\label{sec:data_overview}
We used SkiTB \cite{skitb_a, skitb_b} dataset provided in the competition, this is a single object tracking (SOT) dataset with the aim to track single skier. It provides a comprehensive spatio-temporal video representation and annotation of professional skiing performance. The dataset comes with dense annotations  (ground truth is present for all the frames) for tracking purposes, but it is designed to serve as a well-curated benchmark for subsequent higher-level skiing performance understanding tasks. It covers a wide range of disciplines, athlete styles, courses, and locations, enabling robust testing across diverse skiing conditions. 

\begin{table}[h]
    \centering
    \begin{tabular}{|c|c|c|}
        \hline
        \textbf{Discipline}  & \textbf{No. of Videos} & \textbf{No. of Frames} \\
        \hline
        AL(Alpine skiing)   & 100   & 215517   \\
        \hline
        JP (Jumping skiing)   & 100   & 38201   \\
        \hline
        FS (Freestyle skiing)   & 100   & 99260   \\
        \hline
    \end{tabular}
    \caption{SkiTB data Statistics}
    \label{table:data-stats}

\end{table}
\section{Methodology}
\label{sec:methodology}
Our approach is built upon the \textbf{STARK} \cite{stark} inference pipeline to track a target skier across multi-camera video sequences

\subsection{STARK Tracking Pipeline}

During inference, the pipeline initializes two templates (initial and dynamic) using the ground truth bounding box from the first frame. Template embeddings are extracted after scaling the region by a predefined template factor. For subsequent frames, the model processes a search region cropped based on the previous frame’s bounding box and combines its embeddings with the template embeddings to predict the target’s bounding box and a confidence score. While the initial template remains fixed, the dynamic template is updated when the update interval ($T_u$) has elapsed and the confidence score ($\tau$) exceeds a threshold. If both conditions are met, a new template is extracted from the predicted bounding box and replaces the existing dynamic template, enhancing adaptability to appearance variations and occlusions.

\subsection{Challenges of STARK on SkiTB Dataset}

While STARK \cite{stark} demonstrates strong performance under general tracking conditions, its effectiveness diminishes in the SkiTB \cite{skitb_a, skitb_b} dataset due to challenges introduced by multi-camera transitions. Firstly, sudden camera switches often cause the skier to appear in an unexpected position. Since the tracker determines the search region based on previous frame predictions, it frequently fails to encompass the target, leading to tracking loss. Secondly, rapid skier movement across varying camera perspectives results in abrupt changes in bounding box size and aspect ratio. The fixed search factor struggles to accommodate these variations, causing tracking drift or failure. Furthermore, temporary occlusions, such as obstacles or motion blur, increase the likelihood of the tracker misclassifying background noise as the target or losing the skier entirely. As the search region is determined using past frame references, reidentification becomes difficult when the skier reappears. These tracking errors compound over consecutive frames, further deteriorate the performance. These limitations underscore the necessity of adaptive modifications to enhance tracking robustness and stability.

\subsection{Modifications to the Existing STARK Tracking/Inference Pipeline}

To address the above challenges, we introduced a few key modifications.
 
 \subsubsection{Reattempt with Dynamic Search Factor Adjustment Based on Confidence Score and Object Size}

 To improve target recovery in uncertain frames where confidence scores were low, we implemented a Reattempt mechanism. This involved re-locating the target object by expanding the search area beyond the initial search region, which was originally five times the previous frame’s bounding box size. The search factor for the reattempt was calculated based on the bounding box-to-frame ratio, ensuring that the expanded search region covered up to 60\% of the frame for optimal recovery. The reattempt was triggered when the predicted bounding box was too small or narrow, as the fixed search factor made the subsequent search area insufficient to locate the object. Additionally, the reattempt was activated when the confidence score fell below a critical threshold, indicating uncertainty in detection within the provided search area.

 \subsubsection{Incremental Template Update (ITU) strategy}

 We introduced an Incremental Template Update (ITU) strategy to enhance adaptability. The standard STARK \cite{stark} model updated the dynamic template at fixed intervals, provided the confidence score exceeded a predefined threshold. However, this rigid update strategy often failed to capture key appearance changes between updates. To address this limitation, we modified the update mechanism to allow template updates at any frame where the confidence score surpassed a higher threshold, provided that the last update had occurred at least a certain number of frames ago. These modifications complemented, rather than replaced, the existing update mechanism, preventing template stagnation and enabling more frequent updates when the skier’s appearance changed significantly. As a result, the tracker maintained higher robustness and adaptability across varying conditions.

\section{Implementation details}\label{sec:implementation-details}

\subsection{Data Preparation}

We used 54\% of the data as the train set, 6\% as validation set and the remaining as test set. The dataset provided is imbalanced in terms of the number of frames per discipline, see table~\ref{table:data-stats}, we applied balanced sampling, where sampling weights were assigned based on the inverse occurrence frequency of each discipline. This ensured that underrepresented disciplines contributed equally to a batch of data prepared for training. The sampling weight \(w_i\) was computed as:

\begin{equation}
    w_i = \frac{\max{\left(N\right)}}{N_i}
    \label{eq:weight}
\end{equation}

where \(w_i\) is the sampling weight, \(N_i\) is the number of samples in discipline \(i\), and \(\max{\left(N\right)}\) is the highest sample count among all disciplines. Table~\ref{table:sampling-weights} summarizes the sampling weights used for balanced training across skiing disciplines.

\begin{table}
\centering

\begin{tabular}{| >{\centering\arraybackslash}p{2cm} | >{\centering\arraybackslash}p{2cm} | >{\centering\arraybackslash}p{2cm} |}
\hline
\textbf{Discipline} & \textbf{\#Training samples} & \textbf{Sampling weights}\\
\hline
Alpine (AL) & 114,575
& 1.000 \\
\hline
Freestyle (FS) & 53,389
& 2.146 \\
\hline
Jumping (JP) & 20,536
& 5.579 \\
\hline
\end{tabular}   
\caption{Sampling weights for balanced training across skiing disciplines}
\label{table:sampling-weights}
\end{table}

\subsection{Training Configuration}

Our training process leveraged the MMTracking \cite{mmtrack_lib} library, which provided a modular framework for implementing and modifying tracking models. We used the pretrained STARK \cite{stark} model (STARK-ST2 with ResNet50 as the backbone), which was pre-trained on GOT-10k\cite{got_10k} dataset, and then fine-tuned it on SkiTB \cite{skitb_a, skitb_b} dataset.

Finetuning of the model was carried out in two stages. First, the transformer was fine-tuned for bounding box prediction, ensuring accurate localization of the target skier. The training loss in this stage was a combination of regression and GIoU loss, optimizing the model’s ability to generate precise bounding box coordinates. Second, to improve the confidence scoring of predictions, a scoring head was added on top of the model obtained after the first stage. The model was then further fine-tuned, with only the scoring head being updated while the rest of the model remained frozen. This step enhanced the model’s ability to differentiate between true positives (presence of the skier) and false positives (such as background, occlusions, or other skiers), using cross-entropy loss as the optimization criterion.

Table~\ref{table:training-configurations} summarizes the fine-tuning configurations, while Table~\ref{table:training-hyperparameters} lists the key hyperparameters, determined empirically through grid search and validation performance.

\begin{table}
\centering

\begin{tabular}{|c|c|c|}
\hline
\textbf{Parameter} & \textbf{Stage-1} & \textbf{Stage-2} \\
\hline
Batch Size & 8 & 8 \\
\hline
Learning Rate & 1e-03 & 1e-04 \\
\hline
LR scheduler & \makecell{Step reduction \\ (gamma: 0.75)} & \makecell{Step reduction \\ (gamma: 0.75)} \\
\hline
Optimizer & AdamW & AdamW \\
\hline
Epochs & 100 & 50 \\
\hline
\end{tabular}
\caption{Training/Finetuning configurations for Stage-1 and Stage-2 fine-tuning.}
\label{table:training-configurations}
\end{table}

\begin{table}
\centering

\begin{tabular}{| l  |c|}
\hline
\textbf{Hyperparameter} & \textbf{Value} \\
\hline
Search Factor & 5.0 \\
\hline
Search image size & (320, 320) \\
\hline
Template Factor & 2.0 \\
\hline
Template Size & (128, 128) \\
\hline
Update Interval & 200 frames \\
\hline

\end{tabular}
\caption{Model’s key hyperparameters during finetuning, determined empirically by performing grid search and estimating validation set performance.}
\label{table:training-hyperparameters}
\end{table}

\subsection{Score Head Architecture Modification}

The STARK \cite{stark} model utilizes a score head network with a fixed hidden dimension of 256 for stage-2 training. Instead of maintaining a uniform depth, we introduce a hierarchical structure with varying layer sizes : 512 → 512 → 256 during stage-2 finetuning.
This modification enhances feature extraction in the initial layers while maintaining computational efficiency in later stages. The updated score head improves the model’s ability to distinguish between the positive class (presence of a skier) and the negative class (absence of a skier). This, in turn, improves tracking robustness during inference, as the search factor adjustment is dynamically based on the confidence score of the predicted bounding box.

\subsection{Inference Pipeline Configurations}

Table~\ref{table:inference-hyperparameters} mentions key hyperparameters used during model inferencing.

\begin{table}
\centering

\begin{tabular}{|c|c|}
\hline
\textbf{Parameter} & \textbf{Value} \\
\hline
\makecell{Dynamic Template
Update \\ Interval ($T_u$)}& 200 frames \\
\hline
Confidence Score Threshold ($\tau$) & 0.50 \\
\hline
\makecell{ITU Confidence Score \\ Threshold ($ITU_{\tau}$)} & 0.55 \\
\hline
\makecell{ITU Dynamic Template Update \\ Interval ($ITU_u$)} & 100 frames \\
\hline
\makecell{Low Confidence Threshold \\ for Reattempt(${\tau}_{low}$) }& 0.14 \\
\hline
Bounding Box size for Reattempt & $w \times h \leq 105$ px \\
\hline

\end{tabular}
\caption{Key inference hyperparameters values, determined empirically.}
\label{table:inference-hyperparameters}
\end{table}


\section{Results}
\label{sec:results}
We utilized SkiTB-toolit\footnote{Evaluation Framework - https://github.com/matteo-dunnhofer/SkiTB-toolkit} to evaluate our tracker. To show the effectiveness of our approach we evaluated three version of STARK \cite{stark} model namely, \textbf{STARK}, \textbf{STARK-ski} and \textbf{STARK-ski-ours}.

\textbf{STARK} \cite{stark} model is pretrained on GOT-10K \cite{stark} dataset and is inferenced on SkiTB \cite{skitb_a, skitb_b} dataset with Search Factor of 5 \& Incremental Template Update – ITU of 200. Whereas, \textbf{STARK-ski} is STARK \cite{stark} model finetuned on SkiTB \cite{skitb_a, skitb_b} dataset with non-hierarchical scoring head and inferenced with Search Factor as 5 and ITU as 200. Additionally, \textbf{STARK-ski-ours} is the STARK \cite{stark} model finetuned on SkiTB \cite{skitb_a, skitb_b} dataset with modified model’s scoring head and updated inference pipeline (with Adaptive Search Factor \& Incremental Template Update – ITU).


\begin{table}[ht]
\centering
\begin{tabular}{|p{1.4cm}|p{1.4cm}|>{\centering\arraybackslash}p{1.1cm}|>{\centering\arraybackslash}p{1.1cm}|>{\centering\arraybackslash}p{1.1cm}|}
\hline
\textbf{Discipline} & \textbf{Metric} & \textbf{STARK} & \textbf{STARK-ski} & \textbf{STARK-ski-ours} \\
\hline

All & F1-score & 0.569 & 0.793 & 0.805 \\
 & Precision & 0.580 & 0.797 & 0.805 \\
 & Recall & 0.562 & 0.790 & 0.807 \\
\hline
Alpine & F1-score & 0.529 & 0.822 & 0.834 \\
 & Precision & 0.547 & 0.834 & 0.844 \\
 & Recall & 0.513 & 0.810 & 0.826 \\
\hline
Jumping & F1-score & 0.587 & 0.861 & 0.864 \\
 & Precision & 0.607 & 0.861 & 0.859 \\
 & Recall & 0.572 & 0.862 & 0.870 \\
\hline
Freestyle & F1-score & 0.592 & 0.695 & 0.718 \\
 &  Precision & 0.587 & 0.694 & 0.711 \\
 & Recall & 0.599 & 0.698 & 0.726 \\
\hline
\end{tabular}
\caption{Test set Performance, of different versions of STARK model}
\label{table:test-set-performance}
\end{table}

Results in Table~\ref{table:test-set-performance} show that fine-tuning on domain-specific data significantly enhances tracking performance, while our proposed modifications to the model architecture and inference mechanism further improved performance validating the effectiveness of our proposed adaptive tracking strategies. The improvements are consistent across all skiing disciplines, making the model more robust to camera transitions, occlusions, and scale variations.  However, performance of model was lower in those video sequences of Freestyle (FS) skiing where there were multiple skiers present in the video, causing rapid identity switches between the frames, which lead to poor F1-score for these videos and thus reduced overall FS performance.
\section{Conclusion \& Future Work}
\label{sec:conclusion}
Finetuning on SkiTB \cite{skitb_a, skitb_b} data significantly improves performance compared to using a generic GOT-10k \cite{got_10k} pretrained model.  Additionally, our adaptive modifications (ITU + Dynamic Search Factor) contribute to better tracking stability in multi-camera sequences.

\subsection{Future Enhancements}

Transformer based model are computationally expensive, limiting its use in real-time applications. Future work will explore more computationally efficient architectures without compromising on model’s performance.

We look forward to working in the following directions to improve existing method and explore better computationally efficient alternatives:

\textbf{Refined Confidence Scoring Mechanism}: Improving confidence score estimation to better distinguish between true positives and false positives (e.g., background, occlusions, or other skiers). Aslo, more accurate confidence scores would enable designing better adaptive strategies.

\textbf{Exploring Computationally Efficient Tracking Methodologies}: Investigating Detection-Based Tracking and Joint Detection \& Tracking methods. Additionally, customizing popular and effective tracking algorithms like SMILEtrack \cite{smiletrack} for efficient skier tracking in Tracking by Detection methods.

{\small
\bibliographystyle{ieee_fullname}
\bibliography{egbib}
}
\end{document}